\documentclass{article}

\PassOptionsToPackage{numbers}{natbib}

\usepackage[preprint]{neurips_2023}

\usepackage[utf8]{inputenc} 
\usepackage[T1]{fontenc}    
\usepackage{hyperref}       
\usepackage{url}            
\usepackage{booktabs}       
\usepackage{amsfonts}       
\usepackage{nicefrac}       
\usepackage{microtype}      
\usepackage{xcolor}         
\usepackage{soul}
\usepackage{graphicx}
\usepackage{verbatim}
\usepackage{caption}
\usepackage{subcaption}
\usepackage{footmisc}
\usepackage{booktabs}
\usepackage{hhline}
\usepackage{multirow}
\usepackage{arydshln}
\usepackage{float}
\usepackage{xltabular}
\usepackage{lmodern}
\usepackage[most]{tcolorbox}
\tcbuselibrary{skins}
\usepackage{algpseudocode}
\usepackage{longtable}
\usepackage{framed}
\usepackage{hyperref}
\usepackage{enumitem}  
\usepackage{adjustbox}
\usepackage{makecell}
\usepackage{multirow}
\usepackage{booktabs}
\usepackage{amsmath}  
\usepackage[framemethod=TikZ]{mdframed}  
\usepackage{caption}  
\usepackage{algpseudocode}    
\usepackage{listings} 
\usepackage{booktabs}
\usepackage{makecell}
\usepackage{alltt}
\usepackage{fancyvrb}
\usepackage{ulem}

\tcbset{
  aibox/.style={
    width=\textwidth,
    top=0pt, bottom=0pt, left=5pt, right=5pt,
    colback=white,
    colframe=black,
    colbacktitle=black,
    enhanced,
    center,
    attach boxed title to top left={yshift=-0.1in,xshift=0.15in},
    boxed title style={boxrule=0pt,colframe=white,},
  }
}
\newtcolorbox{AIbox}[2][]{aibox,title=#2,#1}

\usepackage[ruled,vlined]{algorithm2e}
\newcommand{\squishlist}{
   \begin{list}{$\bullet$}
    { \setlength{\itemsep}{0pt}      \setlength{\parsep}{3pt}
      \setlength{\topsep}{3pt}       \setlength{\partopsep}{0pt}
      \setlength{\leftmargin}{1.5em} \setlength{\labelwidth}{1em}
      \setlength{\labelsep}{0.5em} } }

\newcommand{\squishlisttwo}{
   \begin{list}{$\bullet$}
    { \setlength{\itemsep}{0pt}    \setlength{\parsep}{0pt}
      \setlength{\topsep}{0pt}     \setlength{\partopsep}{0pt}
      \setlength{\leftmargin}{2em} \setlength{\labelwidth}{1.5em}
      \setlength{\labelsep}{0.5em} } }

\newcommand{\squishend}{
    \end{list}  }

\usepackage{blindtext}

\newtcolorbox[list inside=mybox,auto counter,number within=section]{MyBox}{colbacktitle=yellow,coltitle=black,title={MyBox \thetcbcounter}}

\newcommand{\sysname}{Orca-3} 
\newcommand{\methodname}{AgentInstruct} 

\usepackage{tikz}

\title{%
\raisebox{-0.3cm}{\includegraphics[width=1cm, height=1cm]{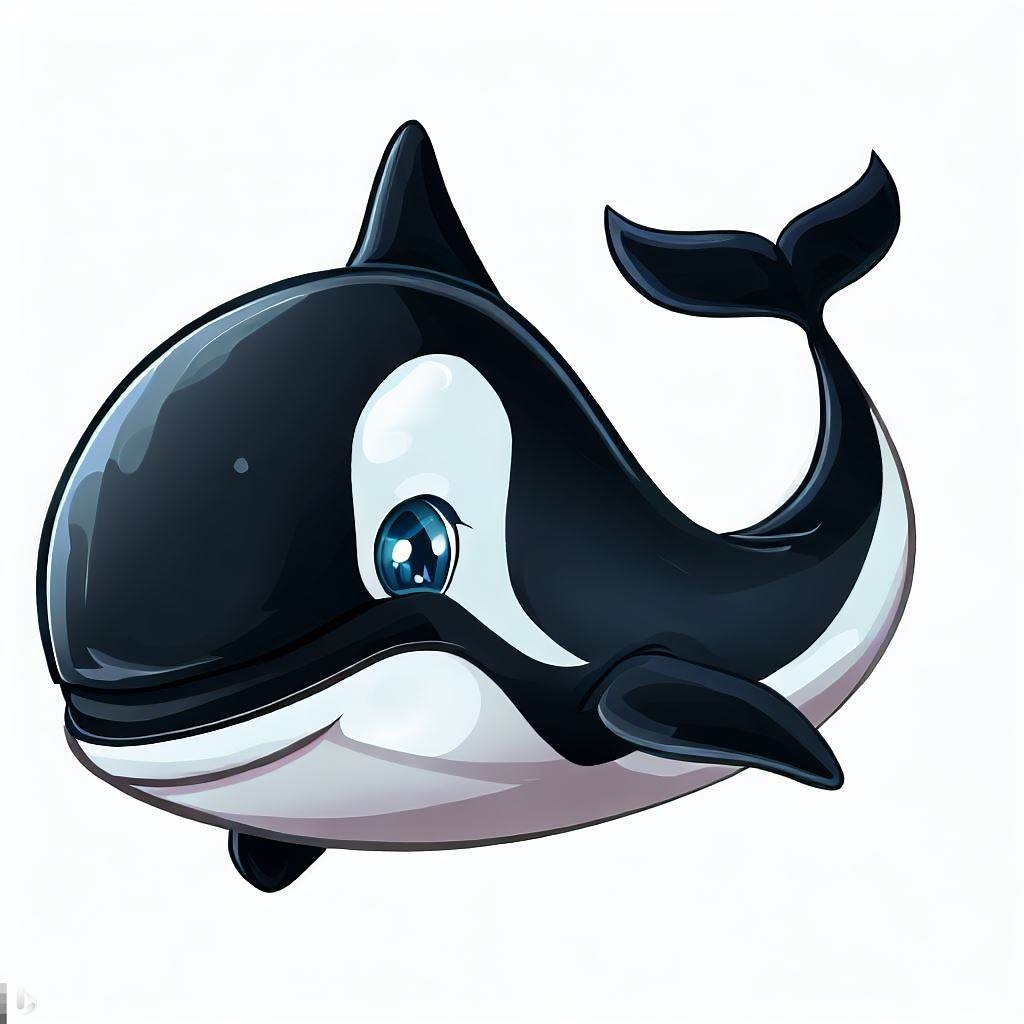}}\ {\methodname:\\ ~Toward Generative Teaching with Agentic Flows}
}

\author{{Arindam Mitra, Luciano Del Corro, Guoqing Zheng, Shweti Mahajan, }\\{ \bf Dany Rouhana, Andres Codas,  Yadong Lu, Wei-ge Chen, Olga Vrousgos,} \\{\bf  Corby Rosset, Fillipe Silva, Hamed Khanpour, Yash Lara, Ahmed Awadallah} \AND Microsoft Research\vspace{0em}}
\begin{document}

\maketitle
\begin{abstract}
Synthetic data is becoming increasingly important for accelerating the development of language models, both large and small. Despite several successful use cases, researchers also raised concerns around model collapse and drawbacks of imitating other models. This discrepancy can be attributed to the fact that synthetic data varies in quality and diversity. Effective use of synthetic data usually requires significant human effort in curating the data.
We focus on using synthetic data for post-training, specifically  creating data by powerful models to teach a new skill or behavior to another model, we refer to this setting as \textit{Generative Teaching}. We introduce AgentInstruct, an extensible agentic framework for automatically creating large amounts of diverse and high-quality synthetic data. AgentInstruct can create both the prompts and responses, using only raw data sources like text documents and code files as seeds. We demonstrate the utility of AgentInstruct by creating a post training dataset of 25M pairs to teach language models different skills, such as text editing, creative writing, tool usage, coding, reading comprehension, etc. The dataset can be used for instruction tuning of any base model. We post-train Mistral-7b with the data. When comparing the resulting model (\sysname) to Mistral-7b-Instruct (which uses the same base model), we observe significant improvements across many benchmarks. For example,  40\% improvement on AGIEval, 19\% improvement on MMLU, 54\% improvement on GSM8K, 38\% improvement on BBH and 45\% improvement on AlpacaEval. Additionally, it consistently outperforms other models such as LLAMA-8B-instruct and GPT-3.5-turbo.

\vspace{-5pt}
 \begin{figure}[ht]
     \centering
     \includegraphics[width=\textwidth]{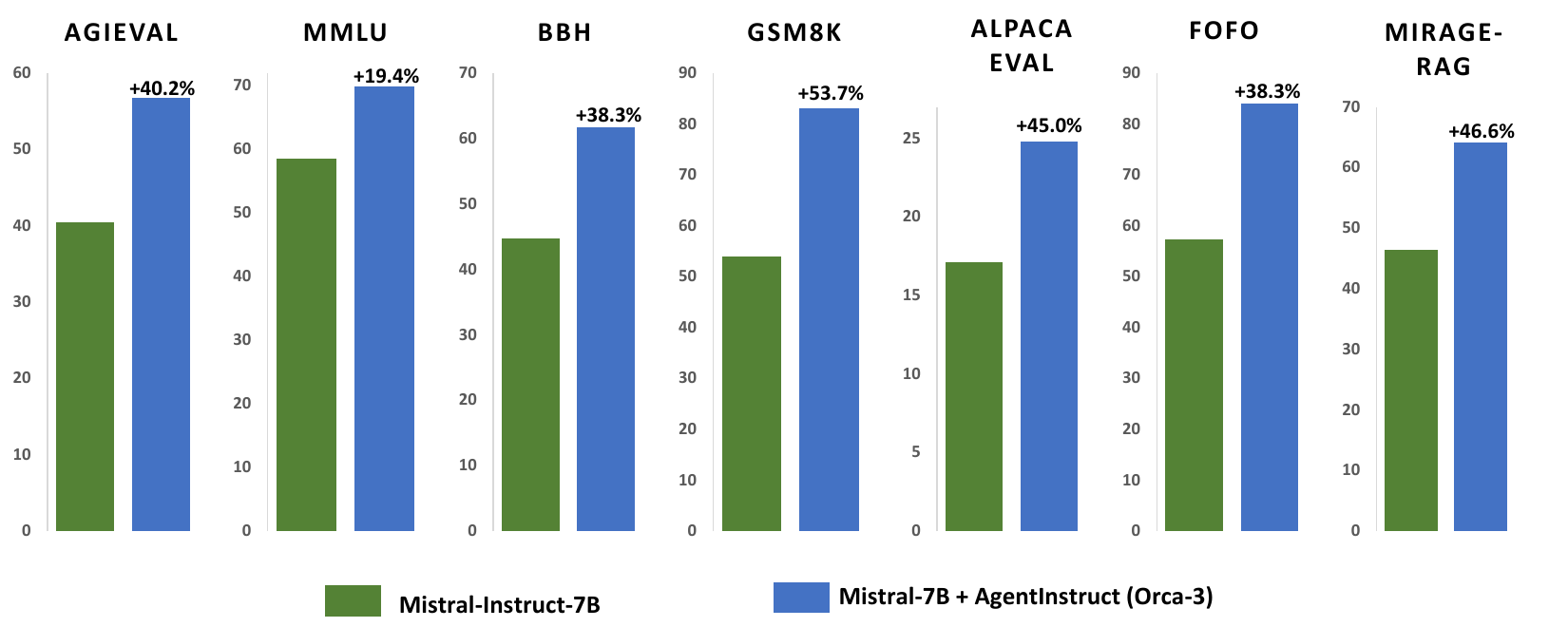}
     \caption{Effect of using AgentInstruct data for post-training Mistral-7B}
     \label{fig:orca3}
 \end{figure}

\end{abstract}
\clearpage

\section{Introduction}

\textbf{Synthetic data accelerated the development of LLMS}: The rise of synthetic data in the training of Large Language Models (LLMs) has been a significant development of the last year. Synthetic data was used to significantly accelerate the progress of model training (especially SLMs) in all stages of training from pre-training (e.g., ~\cite{phi3}), to instruction-tuning (e.g., ~\cite{mukherjee2023orca,xu2023wizardlm}) and RLHF(e.g., ~\cite{lee2023rlaifscalingreinforcementlearning,rosset2024directnashoptimizationteaching}). 
 
\textbf{Generating high quality synthetic data is hard}: On the other hand, research has also shown that pre-training models on synthetic data generated by other models can lead to model collapse~\cite{shumailov2024curserecursiontraininggenerated}, leading to models gradually degenerating as a result. Similar arguments have been made against using synthetic data for pos-training, which could amount to an imitation process that could teach the trained model to pick only stylistic characteristics and not real capabilities~\cite{gudibande2023falsepromiseimitatingproprietary}.
This  discrepancy could be explained by the observation that creating high-quality and diverse synthetic data is hard ~\cite{Cosmopedia}. Successful use of synthetic data involved significant human effort in curating and filtering the data to ensure high quality.
If we focus on post-training synthetic data, we will see the most widely used approach includes starting with a set of prompts and using a powerful model such as GPT-4 \cite{gpt4} to generate responses to these prompts ~\cite{peng2023instructiontuninggpt4} or of an expanded set of the prompts ~\cite{xu2023wizardlm}. This recipe was further improved by eliciting explanations or step-by-step instructions from the teacher model~\cite{mukherjee2020xtremedistil} or using more complex prompting techniques to elicit higher quality answers~\cite{mitra2023orca2teachingsmall}.

\textbf{Synthetic data meets Agents}: 
Another major development we witnessed last year is the rise of Agentic (especially multiagent) workflows ~\cite{wu2023autogenenablingnextgenllm,li2023camelcommunicativeagentsmind}. Agentic workflows can generate high quality data, that surpasses the capabilities of the underlying LLMs, by using flows with reflection and iteration, where agents can look back at solutions, generate critiques and improve solutions. They can also use tools (e.g. search apis, calculator, code interpreters) addressing limitations of LLMs.
Multi-agent workflows bring in additional benefits such as simulating scenarios where we can generate both new prompts and the corresponding responses. They also enable automation of the data generation workflows reducing or eliminating need for human intervention on some tasks.

\textbf{Generative Teaching \& Orca AgentInstruct}: Generating synthetic data for post-training often relies on an existing prompt set that is used as is or used as seeds for generating more instructions.  In this work, we generalize the problem settings to a broader objective of generating abundant amounts of diverse, challenging and high-quality data to teach a particular skill to an AI model, we refer to this setting as Generative Teaching.   AgentInstruct is an agentic solution for Generative Teaching. AgentInstruct focuses on creating demonstration and feedback data and requires only raw documents as input. When generic data is used as seeds, AgentInstruct can be used to teach an LLM a general capability (e.g. Math, Reasoning, RAG, etc.). Domain specific data (e.g. gaming, finance) can also be used as seeds to improve the model in a certain specialization. AgentInstruct  can create:
\begin{enumerate}
    \item High-quality data: using powerful models like GPT-4, coupled with tools like search and code interpreters. 
    \item Diverse data: AgentInstruct generates both prompts and responses. It  uses a large number of agents (equipped with powerful LLMs, tools and reflection flows) and a taxonomy (of over 100 subcategories) to create diverse and high quality prompts and responses,
    \item Large quantities of data: AgentInstruct can run autonomously and can apply flows for verification and data filtering. It does not require seed prompts and uses raw documents for seeding.
\end{enumerate}

Using raw data (unstructured text documents or source code) as seeds has two benefits. First, this data is available in abundance enabling the use of AgentInstruct to create large amounts of diverse data. Additionally, using raw data as seeds, and hence, avoiding using existing prompts, as is or after paraphrasing, can promote learning more general capabilities as opposed to benchmark-specific ones.  

We demonstrate the utility of AgentInstruct by creating a comprehensive synthetic post-training dataset of 25 million prompt and response pairs. The dataset covers a wide array of skills including creative writing, reasoning, math, RAG, tool use, etc. To assess the value of the data, we  use it to finetune Mistral-7B\cite{jiang2023mistral} model. The finetuned Mistral model (Orca-3) shows significant improvement over other instruction-tuned models using the same base model. For example, compared to Mistral-Instruct-7B, it shows 40\% improvement on AGIEval, 19\% improvement on MMLU, 54\% improvement on GSM8K, 38\% improvement on BBH,  45\% improvement on AlpacaEval and 31.34\% reduction on hallucination across multiple summarization benchmarks. Additionally, it  outperforms other models such as LLAMA-8B-instruct and GPT-3.5 on multiple benchmarks. Note that the only seed data used is publicly available raw materials and no task-specific or benchmark data has been used as seeds.

While we demonstrate the utility of AgentInstruct by creating a generic post-training synthetic dataset, we believe that agents can enable the creation of Synthetic-Data-Generation-As-A-Service where we start with raw materials (e.g. web data for general model training or domain specific data for specialized models), and we generate data for post-training and finetuning, hence enabling continual learning and improvement of any base LLM. Additionally, we believe that the AgentInstruct approach can be used for self-improvement of larger, more capable models because of: (1) the ability to generate new prompts and (2) the ability to generate responses that exceed the quality of the LLM used in the agentic flow (because of the use of tools, reflection, etc.).

\section{Generative Teaching: AgentInstruct}

Creating synthetic datasets for supervised fine-tuning and instruction-tuning has seen significant progress over the last year. The quality of these datasets has been steadily improving. High quality can be achieved by using powerful frontier models (or agenetic flows based on these models) to generate responses. However, when creating synthetic data, in addition to quality, we also need to consider several other fundamental questions:  
  
\begin{enumerate}  
  \item How can we create a vast amount of data?  
  \item How can we ensure that the generated data is diverse?  
  \item How can we generate complex or nuanced data points?  
\end{enumerate}  
  
In the AgentInstruct methodology, we outline a structured approach to tackle these challenges as follows:
  
\begin{mdframed}[innertopmargin=10pt, innerbottommargin=10pt, innerleftmargin=10pt, innerrightmargin=10pt, backgroundcolor=gray!10, roundcorner=10pt]    
\captionof{figure}{Concise Summary of the AgentInstruct Methodology}    
\begin{lstlisting}[mathescape=true, breaklines=true, columns=fullflexible]    
1. Assemble a collection of raw seeds (e.g., textbook chapters, web articles, code snippets).    
2. $\textbf{for}$ each seed in the collection $\textbf{do}$     
3. Transform the seed with the aid of one or more  content transformation Agents (Content Transformation Flow).  
4.  Route it through a series of instruction creation Agents to create a diverse set of instructions (Seed Instruction Creation Flow).  
5.  Utilize another group of Refinement Agents to iteratively refine the complexity and quality of the seed instructions (Refinement Flow).  
6. $\textbf{end for}$    
\end{lstlisting}  
\end{mdframed}  
  
We use agentic flows to automate the generation process and leverage raw articles as seeds to foster diversity and ensure that problems generated in different iterations (line 2 in Figure 1) are distinct and of broad coverage. This enables us to create data at scale (benefiting from automation of agentic flows), with high diversity (based on the broad and diverse seeds) and of varying complexity (benefiting from the iterative and refinement patterns supported by agentic flows). AgentInstruct defines three different flows:
  
\textbf{Content Transformation Flow} converts the raw seed into an intermediate representation that simplifies the creation of instructions tailored to specific objectives. It comprises of multiple agents and is often instrumental in the generation of high-quality data and serve as an additional means to introduce diversity. 

\textbf{Seed Instruction Generation Flow} comprising of multiple agents takes as input the transformed seed from the Content Transformation Flow and generates a set of diverse instructions. The only goal of the Seed Instruction Flow is to introduce diversity for which it often relies on a pre-defined, but extensible, taxonomy.

\textbf{Instruction Refinement Flow} takes as input the instructions from the Seed Instruction Flow and iteratively enhances their complexity and quality. Towards this we use the concept of Suggester-Editor Agents\cite{mitra2024orca}. Suggester agents initially propose various approaches to increase the intricacy of the initial instructions (making them more complex, unsolvable, or tricky), after which the Editor agents modify the instructions in accordance with these suggestions.

Each flow consists of a number of agents. We use a generic definition of an agent, where an agent is powered by an LLM and can optionally have the ability to use tools such as search APIs, code interpreter or a calculator. Each agent has a specific role and set of instructions specified as part of the underlying LLM system message.

\begin{figure}[t]
    \centering
    \includegraphics[width=\textwidth]{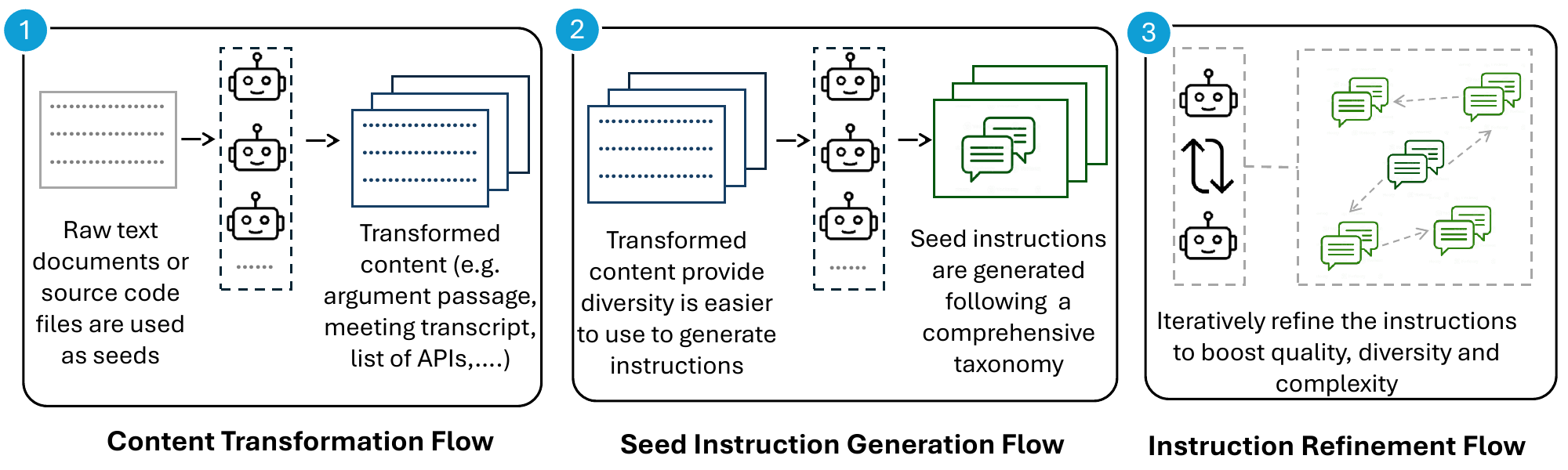}
    \caption{Figure provides a thematic overview of the roles played by different groups of agents. Content Transformation Flow converts the seed into an intermediate representation that makes it easier to create high quality and diverse data. Seed Instruction Generation Flow creates instances of the target tasks following a taxonomy. Refinement Flow explores the space further by starting from these initial data points and exploring the neighborhood. The expectation is that by picking a random seed we will be able to cover the entire region of data points.}
    \label{fig:ag}
\end{figure}

We implemented these flows for 17 different skills, each having multiple subcategories. The skills include reading comprehension, question answering, coding, retrieval augmented generation, creative writing, tool/API use and Web control. The full list is provided in Table~\ref{tab:AgentInstruct_caps}

We explain how the workflows work with case studies of generating data for the following three skills:

\begin{description}  
  \item[Reading Comprehension:] The ability to understand, process, and interpret written text.  
  \item[Text Modification:] The process of altering text to suit different purposes.  
  \item[Tool Use:] The employment of functions or APIs to perform tasks or solve problems.  
\end{description} 

\begin{table}[t]
\centering
\small
\begin{tabular}{p{\linewidth}}
\toprule
\textbf{Reading Comprehension:}Reading comprehension is a critical skill involving processing and understanding text, which is necessary for learning and encompasses decoding, fluency, and vocabulary knowledge. Reading comprehension tests typically present text passages of varying lengths and subjects, followed by questions that assess the reader’s understanding.    \\
\midrule
\textbf{Open Domain Question Answering:} Open domain question answering involves generating responses to questions over a wide range of topics, without being restricted to a specific domain. \\
\midrule
\textbf{Text Modification:} Text modification involves changing existing text to improve its quality, modify its tone, or to fit a specific context or audience. It is a common task in content creation and editing. \\
\midrule
\textbf{Web Agent:} A web agent is a software program that autonomously performs tasks on the web, such as where to click, how much to scroll.  \\
\midrule   
\textbf{Brain Teaser} A brain teaser is a problem or puzzle, typically requiring thought to solve, often for amusement but also used for training logical thinking and problem-solving skills. \\
\midrule
\textbf{Analytical Reasoning} Analytical reasoning involves the ability to look at information, be it qualitative or quantitative in nature, and discern patterns within the information. It's a process that includes understanding a system of relationships and draw logical conclusions about those relationships. \\
\midrule
 \textbf{Multiple Choice Questions} Multiple choice questions are a form of assessment where respondents are asked to select the best possible answer (or answers) out of the choices from a list. They are common in standardized tests, quizzes, and surveys.  \\
\midrule   
 \textbf{Data To Text} Data-to-text refers to human-readable textual summaries from source data. These could be used for generating reports, explanations, or narratives from structured data.  \\
\midrule   
 \textbf{Fermi} Fermi problems are estimation problems which seek quick, rough estimates of quantities which can be difficult to measure. Named after physicist Enrico Fermi, these problems often require making justified guesses or assumptions to reach a solution.  \\
\midrule   
 \textbf{Coding} Coding involves writing code following instructions, understanding code, debugging code, tracing or writing test cases. \\
\midrule   
 \textbf{Text Extraction} Text extraction is the process of retrieving relevant information from a larger text document. This can include tasks like named entity recognition, keyword extraction, or extracting specific data fields from unstructured text. \\ 
\midrule   
 \textbf{Text Classification} Text classification is a type of machine learning task where text documents are automatically classified into predefined categories. This can be used for spam detection, sentiment analysis, and topic labelling among others.  \\
\midrule   
 \textbf{Retrieval Augmented Generation} Retrieval Augmented Generation (RAG) is a method used in natural language processing that combines retrieval-based and generative models to generate responses. It first retrieves relevant documents and then uses these documents to generate a response. \\  
\midrule   
 \textbf{Tool Use} Tool use involves the manipulation of tools to achieve goals. In AI, this refers to the ability of an AI system to use available resources or auxiliary systems to solve complex tasks. \\  
\midrule   
 \textbf{Creative Content Generation} Creative content generation involves the creation of original content, often involving elements of novelty, value, and surprise. In AI, this could refer to generating text, music, or images that are not only new but also meaningful and interesting. \\  
\midrule   
 \textbf{Few Shot Reasoning} Few-shot reasoning refers to the ability of a machine learning model to understand new concepts, patterns, or tasks with minimal examples or guidance. It's a desired trait in AI, mimicking the human ability to learn quickly from few examples.  \\
\midrule   
 \textbf{Conversation} Conversation refers to conversational agents or chatbots that interact with humans in a natural, human-like manner. \\

\bottomrule
\end{tabular}
\caption{List of 17 capabilities for which we implemented AgentInstruct Flows}
\label{tab:AgentInstruct_caps}
\end{table}

\subsection{AgentInstruct Flow for Reading Comprehension}
Reading comprehension is a critical skill involving processing and understanding text, which is necessary for learning and encompasses decoding, fluency, and vocabulary knowledge. 

Reading comprehension questions range from asking for explicit information (literal comprehension) to requiring inferences, understanding vocabulary in context, analyzing text structure and argumentation, critically evaluating the content, and synthesizing information from different parts of or multiple texts. Reading comprehension is a very important capability and can enable scenarios like question answering, search, grounded reasoning, etc. 

\textbf{Content Transformation Flow}
Web crawls encompass an extensive collection of human-generated text, which holds potential for generating reading comprehension materials. However, these sources are not inherently structured to facilitate the teaching of reading comprehension skills. Consequently, they do not support the consistent generation of diverse question types required for comprehensive reading comprehension evaluation. For instance, the LSAT Logical Reasoning test features specialized question categories, including assumption, strengthening/weakening, flaw, and inference questions. Crafting such questions necessitates passages with a particular stylistic and logical framework. The objective of Content Transformation Flow is to transform arbitrary articles into well-crafted pieces that are conducive to the formulation of a wide array of reading comprehension question types.

Our current flow for Reading Comprehension encompasses a suite of nine content transformation agents for generating argument passages, debates and conversations, long passages, meeting transcripts, poems, satarical content, etc. Detailed description is provided in Appendix~\ref{sec:appendix-additional-details}. 
Given a seed article, the flow will randomly pick one of the Content Transformation Agents to assess the seed article and attempt to generate the text passages. The following provides an example for the \textit{Argument Passage Generator}.

\begin{tcolorbox}
\small
\fbox{EXAMPLE: Content Transformation Flow}\\\\
\textbf{Random Seed}\\
Uric acid is a substance produced naturally by the breakdown of purine (a type of dietary protein). When it is in excess in the body, crystals composed of these substances are formed.  
These crystals are deposited in various parts of the body, mainly in the joints and kidneys, causing pain and other aggravations.  
The lack or excess of uric acid in the body is caused by some diseases (such as leukemia, obesity, kidney diseases, and anemia) and factors related to lifestyle (consumption of alcohol and processed foods, for example).  
  
Contents
\begin{itemize}  
    \item Where does purine come from?  
    \item Where is uric acid found?  
    \item What is high uric acid?  
    \item What can high uric acid cause?  
    \item What is low uric acid?  
    \item Tests: how to know uric acid levels?  
\end{itemize}  
  
Where does purine come from?\\  
Purine is produced and released into the bloodstream when an amino acid is broken down by digestion.  
Therefore, it is produced naturally in the body.  
... (Truncated)  
\end{tcolorbox}

\begin{tcolorbox}
\small
\textbf{Transformed Seed (Argument Passage Generator)}\\
Recent studies have indicated that high levels of uric acid in the blood, known as hyperuricemia, may increase the risk of cardiovascular disease. Uric acid is a byproduct of purine metabolism, and while it is naturally produced in the body, excessive amounts can lead to health complications. Hyperuricemia is typically defined as levels above 6 mg/dL in women and 7 mg/dL in men. This condition can result from increased production of uric acid or insufficient elimination through urine. Factors contributing to high uric acid include a diet rich in purines, found in red meat and seafood, as well as lifestyle choices such as alcohol consumption and physical inactivity. Conversely, low levels of uric acid, or hypouricemia, are less common and usually do not present symptoms. Nonetheless, they can be indicative of underlying kidney or liver issues. Diagnosing either condition requires laboratory blood and urine tests. It is important to note that while high uric acid levels are associated with an increased risk of cardiovascular disease, the causal relationship remains to be conclusively established.
\end{tcolorbox}

\textbf{Seed Instruction Generation Flow}
We have currently compiled a collection of $43$ reading comprehension questions types. The list includes literal comprehension questions, critical comprehension questions, evaluative comprehension questions, reasoning, identifying assumptions, identifying information that strengthens/weakens an argument, ordering events, etc.

A list some of these types is in Appendix~\ref{sec:appendix-additional-details}.

 We have defined multiple Agents targeting these categories. Each of these agent receives as input a piece of text and generates a list of questions based on a predefined question type. The orchestration process will engage a subset of these agents, determined by the content transformation agent in the preceding step. Upon completion of this phase, we will obtain a collection of (passage, question) pairs that will constitute the input for the subsequent stage. The following figure provides an example of \textit{Strengthen} type question that is generated from the running example.

\begin{tcolorbox}
\small
\fbox{EXAMPLE: Seed Instruction}\\\\
Which one of the following, if true, most strengthens the argument that high levels of uric acid increase the risk of cardiovascular disease?\\
(A) Some individuals with hyperuricemia do not develop cardiovascular disease. \\
(B) Physical activity has been found to have no significant impact on uric acid levels in the blood.\\
(C) Dietary supplements have been shown to reduce uric acid levels in the blood.\\
(D) A significant number of patients with cardiovascular disease have been found to have uric acid levels above the defined thresholds for hyperuricemia.\\
(E) Individuals with a family history of cardiovascular disease are more likely to have high uric acid levels.  
\end{tcolorbox}

\textbf{Instruction Refinement Flow}
The refinement flow contains multiple suggester-editor agents that will go through each of the (passage, question) pairs  and create more such pairs with the following goals: (1) Modify the passage to make the question unanswerable, (2) Modify the passage to alter the answer, if possible, in opposite direction, or (3) Modify the questions or answer choices (if applicable) to make them complex.

Here is an example where the suggester agent that tries to add a difficult distractor to confuse the test taker.

\begin{tcolorbox}
\small

   \fbox{EXAMPLE: Refinement Flow}\\\\
\textbf{SUGGESTER}\\
\begin{itemize}
    \item \textbf{Suggestion 1} Introduce a hypothetical study or finding that could potentially strengthen the argument, requiring the test-taker to infer its impact on the relationship between uric acid levels and cardiovascular disease.
    \item \textbf{Suggestion 2} Add a layer of complexity by suggesting a genetic predisposition to hyperuricemia and its correlation with increased cardiovascular events, which would require the test-taker to consider both genetic and physiological factors.
    \item  \textbf{Suggestion 3} Include a distractor option that seems to strengthen the argument but upon closer examination, it does not directly relate to the causal relationship between uric acid levels and cardiovascular disease, thus testing the test-taker's ability to discern relevant from irrelevant information.
\end{itemize}

\textbf{EDITOR}\\
\begin{itemize}
\item \textbf{Modification 1,2} (omitted for brevity)
\item \textbf{Modification 3} Which one of the following, if true, most strengthens the argument that high levels of uric acid increase the risk of cardiovascular disease? \\(A) Some individuals with hyperuricemia do not develop cardiovascular disease.\\ (B) Physical activity has been found to have no significant impact on uric acid levels in the blood.\\ (C) A placebo-controlled trial shows that individuals taking uric acid-lowering medication do not exhibit a reduced incidence of cardiovascular disease compared to the placebo group.\\ (D) A significant number of patients with cardiovascular disease have been found to have uric acid levels above the defined thresholds for hyperuricemia.\\ (E) A cross-sectional study finds that high uric acid levels are associated with other risk factors for cardiovascular disease, such as hypertension and diabetes, but does not establish a direct causal link.
\end{itemize}
\end{tcolorbox}

\subsection{AgentInstruct Flow for Text Modification}
Text modification is the process of editing and refining written content to enhance its quality and effectiveness or alter its attributes. This involves correcting spelling and grammar, clarifying ideas, reorganizing content for better flow, adjusting tone, ensuring style consistency, fact-checking, removing redundancies, formatting, developing content, and adapting to specific audiences. It is a useful skill for LLMs that help in content writing. While several content transformation agents can be introduced to intentionally modify the text as described earlier, we focus here on showing the how the instructions are created and refined.

\textbf{Seed Instruction Generation Flow}  We have currently compiled a collection of $18$ types of text modification tasks including paraphrasing, expansion, simplification, redacting or removing content, styling, code switching, etc. The full list is in Appendix~\ref{sec:appendix-additional-details}.

We define an Agent for each of the task type. Each agent takes as input a piece of text and creates several text modification tasks of the associated type. Here we provide an example input and a task created by the \textit{Paraphrasing} Agent.

\begin{tcolorbox}
\small
    \fbox{EXAMPLE: Seed Instruction}
    
    \textbf{Random Seed}
    
    April 6-8, 2017, University of Iowa, Iowa City, USA.
    Abstracts due December 1, 2016.
    
Finance is hard to escape. In recent years, the increasing social impact and interconnection of ﬁnancial discourses, markets, actors, and institutions have been understood under the broad concept of financialization. Natascha van der Zwan identifies three distinct research streams that have approached financialization as 1) a regime of accumulation, 2) the influence of financial markets and instruments on non-financial corporations as well as the banking and finance industry, and 3) a discourse of risk-taking, self-management and self-fulfillment that is transforming people into investing subjects. Some anthropological skeptics, however, argue that finance has a far longer genealogy than the financialization literature has to date recognized. For example, in the context of a lengthy human history of creating hierarchy, financialization may simply be a new technology serving an old purpose.

... (omitted for brevity)...
· The supply chains of financial products connect different places and political projects across the globe. How do such financial instruments transform social life?

Abstract deadline is December 1, 2016.

1. Go to americananthro.org and log in. If you don’t have a login id and password, create one (you do not need to join the American Anthropological Association).

2. Once you are logged in, look to the left hand column, click on Meeting registration.

3. Click on register under the SEA 2017 Annual Meeting then follow online prompts to register for the meeting (if we do not accept your abstract and you decide not to attend, you may request that your registration fee be refunded and we would be happy to do so).

4. Once you are registered, AAA will automatically send you an email inviting you to submit an abstract. Click the link and follow the instructions.

\textbf{Task:}

Rewrite the event details (date, location, abstract deadline) in a more casual tone.
\end{tcolorbox}

Given a random seed, the seed instruction creation flow randomly picks one of the $18$ agents and uses that to obtain a seed (text, text modification instruction) pair.

\textbf{Instruction Refinement Flow} The refinement flow consists of only one Suggester-Editor pair. The Suggester-Editor duo increase the complexity of the generated instructions while taking as input the text and the task modification instruction as input. Here we show three suggestions and edits as proposed by the suggester-editor pair.

\begin{tcolorbox}
\small

\fbox{EXAMPLE: IRefinement Flow}

\textbf{Suggestions}

\begin{enumerate}[label=\textbf{Suggestion \arabic*:},leftmargin=*]  
  \item Incorporate a fictional narrative. Use a conversational style with colloquial language and include a humorous element.  
  \item Translate the event details into a poetic format. Maintain accurate information while using rhyming couplets and ensure the tone remains light and engaging.  
  \item Frame the event details as a social media post. Use internet slang and emojis. Keep the message within 280 characters.  
\end{enumerate} 

\textbf{Modified Instructions}
\begin{enumerate}[label=\textbf{Instruction \arabic*:},leftmargin=*]  
  
  \item Rewrite the event details (date, location, abstract deadline) as if you're telling a funny story to a friend using casual and colloquial language, while incorporating a fictional narrative that still conveys the necessary information.  
    
  \item Transform the event details (date, location, abstract deadline) into a light-hearted poem with rhyming couplets, ensuring that the essential information is accurately conveyed in a poetic format.  
    
  \item Craft a social media post that includes the event details (date, location, abstract deadline) using internet slang, emojis, and a casual tone, while keeping the message concise and within 280 characters.  
    
\end{enumerate} 
\end{tcolorbox}

\subsection{AgentInstruct Flow for Tool Use}
The task of tool use or API use for LLMs involves enabling models to interact with external tools or services; via APIs. This capability allows AI systems to extend their functionality, access external data, and perform actions beyond their native capabilities. 

\textbf{Content Transformation Flow} We use source code snippets or an API description \cite{qin2023toolllm} as the random seed. If source code snippets has been used as the seed, a content transformation agent is used to synthesize an API description from the code snippet. The goal of the Content Transformation Flow is to synthesize list of APIs from the random seed. API lists are created by either: (1) using an API retrieval agent that iteratively searches for similar code to expand the API list or (2) the agent uses the LLM to hypothesize other APIs present in the library.

The following figure provides an example of the library reconstruction scenario.
 \begin{tcolorbox}
\small
   \fbox{EXAMPLE: Content Transformation Flow [Library Reconstruction]}\\\\
   \textbf{Seed}
    \begin{lstlisting}[breaklines=True, columns=fullflexible]
  {
    "name": "View All Food Items",
    "description": "The request enables clients to obtain a detailed list of food items, complete with nutritional profiles such as calorie count, protein, fat, ...",
    "parameters": {
      "type": "object",
      "properties": {
        "limit": {
          "type": "NUMBER",
          "description": "limit the length of response"
        }
      },
      "required": []
    }
  }
   \end{lstlisting} 
    
   \textbf{Reconstructed Remaining APIs from the Library}
   \begin{lstlisting}[breaklines=true, columns=fullflexible] 
[
  {
    "name": "Search Food Items",
    "description": "Allows clients to search for food items by name and retrieve a list of matching items ...",
    "parameters": {
      "type": "object",
      "properties": {
        "query": {
          "type": "string",
          "description": "The name or partial name of the food item to search for."
        },
        "limit": {
          "type": "number",
          "description": "Optional parameter to limit the number of search results returned."
        }
      },
      "required": [
        "query"
      ]
    }
  },
  \end{lstlisting}
\end{tcolorbox}

\begin{tcolorbox}
\small
\begin{lstlisting}[breaklines=true,  columns=fullflexible] 
  {
    "name": "Get Food Item Details",
    "description": "...",
    "parameters": {
      ...
    }
  },
  {
    "name": "Create Meal Plan",
    "description": "...",
    "parameters": {
       ...
     } 
  },
  {
    "name": "Update Food Item",
    "description": "...",
    "parameters": {
      ...
    }
  },
  {
    "name": "Track User Meal",
    "description": "...",
    "parameters": {
      ...
    }
  },
  {
    "name": "Get Dietary Recommendations",
    "description": "...",
    "parameters": {
      ...
    }
  },
  {
    "name": "Add New Food Item",
    "description": "...",
    "parameters": {
     ...
    }
  },
  {
    "name": "Delete Food Item",
    "description": "...",
    "parameters": {
      ...
    }
  },
  {
    "name": "Get User Nutritional Stats",
    "description": "...",
    "parameters": {
     ...
    }
  }
]
   \end{lstlisting}
\end{tcolorbox}

\textbf{Seed Instruction Creation Flow}
The seed instruction creation flow, consume the list of APIs and employs variety of agents to create several tasks of the following types:
\begin{enumerate}  
    \item Tasks that require the use of a single API:  
    \begin{enumerate}  
        \item Tasks where the input supplies all necessary parameters.  
        \item Tasks where the input includes superfluous parameters.  
        \item Tasks where the input is missing some required parameters.  
    \end{enumerate}  
      
    \item Tasks that necessitate the use of multiple APIs:  
    \begin{enumerate}  
        \item Tasks where the input provides all necessary parameters.  
        \item Tasks where the input is missing some required parameters.  
    \end{enumerate}  
      
    \item Tasks that require a single API which is not listed among the available APIs.  
      
    \item Tasks that require multiple APIs but lack some of the essential APIs in the provided list.  
\end{enumerate} 
\textbf{Refinement Flow}
The goal of the Refinement flow is to increase the complexity, for e.g., looking at the task and the conversation and to suggest refinements to increase the number of steps required to solve the task.

The following figure provides an example of a multi-turn conversation created by the Agent-Instruct flow.

\begin{tcolorbox}
    \small
   \fbox{EXAMPLE: Instruction Data Created by AgentInstruct]}\\\\
   \textbf{System Message}
    
    As an AI assistant, your role is to assist users in achieving their desired outcomes. You have the capability to utilize a variety of APIs, and at each step of the process, you are required to produce a markdown output.  
   
At any given moment, you have two choices to proceed:  
   
Choice 1: Initiate an API Call  
If you choose to call an API, you must generate a makdown including the "name" of the API and the necessary "parameters" for the call. After creating the response, you must wait for the system to execute the API with the provided parameters and return the results to you. The markdown you output should follow this structure:  
\begin{verbatim}
ACTION_TYPE : API_CALL
ACTION: name(parameter_name=value, another_parameter_name=value, ....) 
\end{verbatim}
Choice 2: Conclude the Process  
There are times when you may decide to end the process, either because 
a) you have successfully obtained the answer\\
b) you have determined that the appropriate APIs are not available.\\ 
c) you discovered that user did not provide values for some required parameter \\

In case you successfully obtained the answer use FINAL\_ANSWER as the action name with only one parameter being the final answer. In case you decided to stop due to other reasons, use FAILED as the action with one parameter describing the reason you decided to stop. Overall you should follow this structure:  
\begin{verbatim}
ACTION_TYPE : STOP
ACTION: [ACTION_NAME]([VALUE])  
\end{verbatim}

Below is a list of APIs that you have access to: 
\begin{verbatim}
[
 APIs from the Library construction example both the seed and the 
 generated ones
]  
\end{verbatim}

\textbf{USER:}\\
I want to start a new diet plan for the next week. My goal is to consume around 1500 calories per day, and I prefer vegetarian meals. I'll be eating three meals a day. Can you create a meal plan for me? Also, I'd like to track what I eat every day and get some new food recommendations. At the end of the week, I want to see my nutritional summary. Oh, and I found this great recipe for a Quinoa Salad that I want to add to your database. I also noticed that the calorie count for `Chana Masala' seems off; it should be lower. And please remove 'Butter Chicken' from my list; I've stopped eating it.
\end{tcolorbox}

\begin{tcolorbox}
\small    
\textbf{ASSISTANT:}\\
To assist you with your request, I will perform the following actions:

1. Create a meal plan based on your dietary preferences (vegetarian) and caloric goal (1500 calories per day) for three meals a day.\\
2. Provide you with a way to track your daily meals.\\
3. Offer you new food recommendations.\\
4. Generate a nutritional summary at the end of the week.
5. Add the Quinoa Salad recipe to the database.\\
6. Update the nutritional information for 'Chana Masala'.\\
7. Remove 'Butter Chicken' from the database.\\

Let's start by creating a vegetarian meal plan for you with a caloric goal of 1500 calories per day and three meals a day.

\begin{lstlisting}[breaklines=True]
 ```markdown
ACTION_TYPE : API_CALL
ACTION: Create Meal Plan(dietary_preferences=["vegetarian"], caloric_goal=1500, num_meals=3)
```     
\end{lstlisting}
\textbf{USER:}
\begin{lstlisting}[breaklines=True, columns=fullflexible]
{
  "status": "success",
  "meal_plan": {
    "day_1": {
      "breakfast": {
        "food_items": ["Oatmeal with Fruits", "Almond Milk"],
        "total_calories": 350
      },
      "lunch": {
        "food_items": ["Chickpea Salad", "Whole Wheat Bread"],
        "total_calories": 500
      },
      "dinner": {
        "food_items": ["Mixed Vegetable Stir Fry", "Brown Rice"],
        "total_calories": 650
      }
    },
    "day_2": {
      ...
    },
    ...
  }
}
\end{lstlisting}

\textbf{ASSISTANT:}\\
Your meal plan for the next week has been successfully created. Here's a brief overview:

- **Day 1**: Oatmeal with fruits and almond milk for breakfast, chickpea salad with whole wheat bread for lunch, and mixed vegetable stir fry with brown rice for dinner.\\
- **Day 2**: ....
....

Next, let's proceed to add the Quinoa Salad recipe to the database. Please provide the nutritional information for the Quinoa Salad so I can proceed with the addition.

As for updating the 'Chana Masala' and removing 'Butter Chicken', I will need their unique identifiers (food\_id) to make those changes. If you have those identifiers, please provide them. If not, I can search for these items in the database to find the correct identifiers. Would you like me to search for 'Chana Masala' and 'Butter Chicken'?
\\\par

\textbf{USER:}\\
...(truncated)
\end{tcolorbox}

\section{Orca-3}
\subsection{Dataset Description}

We implemented an AgentInstruct Flow for 17 different capabilities as described in Table~\ref{tab:AgentInstruct_caps}.

We created a collection of approximately 22 million instructions aimed at teaching the aforementioned skills. We have used unstructured text and code files sampled from KnowledgePile\cite{fei2024query}, AutoMathText\cite{zhang2024automathtext}, a subset of openstax  and  a subset of apache-2.0 licensed source code files from~\cite{githubcodeclean}. The dataset covers variety of skills, as detailed in Table~\ref{tab:AgentInstruct_caps}. Using unstructured content as seeds for instruction data generation has several benefits. First, there is abundance of such data enabling the generation of large-scale and diverse instruction data. Additionally, it enables us to avoid using any benchmark-specific data as seeds and hence focus on optimizing for a capability, not for a specific benchmark.

In addition to the 22 million instructions, we have incorporated approximately 3.8 million paired instructions sourced from Orca-1\cite{mukherjee2023orca}, Orca-2\cite{mitra2023orca2teachingsmall}, Orca-Math\cite{mitra2024orca} and samples from other publicly available sources such as ~\cite{ding2023enhancing, yu2023metamath,ivison2023camelschangingclimateenhancing,siddiq2024regexeval}. We refer to this data as Orca-2.5-dataset.

The culmination of these datasets results in approximately 25.8 million paired instructions, all of which are incorporated into the training of Orca-3. Furthermore, we have trained a separate model, referred to as Orca-2.5, using the  3.8 million instructions (Orca-2.5-dataset). The purpose of this is to compare and evaluate the impact of the 22 million instructions curated through AgentInstruct.

\subsection{Training Details}

We use the 25.8 million paired instructions described earlier to finetune Mistral-7b-v0.1. We choose this model because the it makes the weights publicly available for the base (no-instruction-tuned) version, with a permissive license allowing easy redistribution. We refer to the finetuned model ( Mistral-7b-v0.1 finetuned on AgentInstruct dataset) as Orca-3.

Each pair in the dataset undergoes a tokenization process using the Mistral tokenizer, ensuring a maximum sequence length of 8192 with packing. To guarantee that the training loss is calculated based only on the response conditioned on the prompt,  label masking is  applied.  Weight decay was set at 0.1  
  
The finetuneing used 19 NVIDIA A100 nodes, or 152 NVIDIA A100 GPUs, each with a batch size of 10. We used AdamW optimizer with an initial learning rate of 8e-6   and a a cosine learning rate schedule. We also used a linear learning rate warm-up during the initial 500 steps. The model was trained for  three epochs and the training process concluded after approximately 200 hours.

\section{Evaluation Results}
\subsection{Orca-Bench}
\begin{figure}[!htb]  
    \centering  
    \includegraphics[scale=0.5]{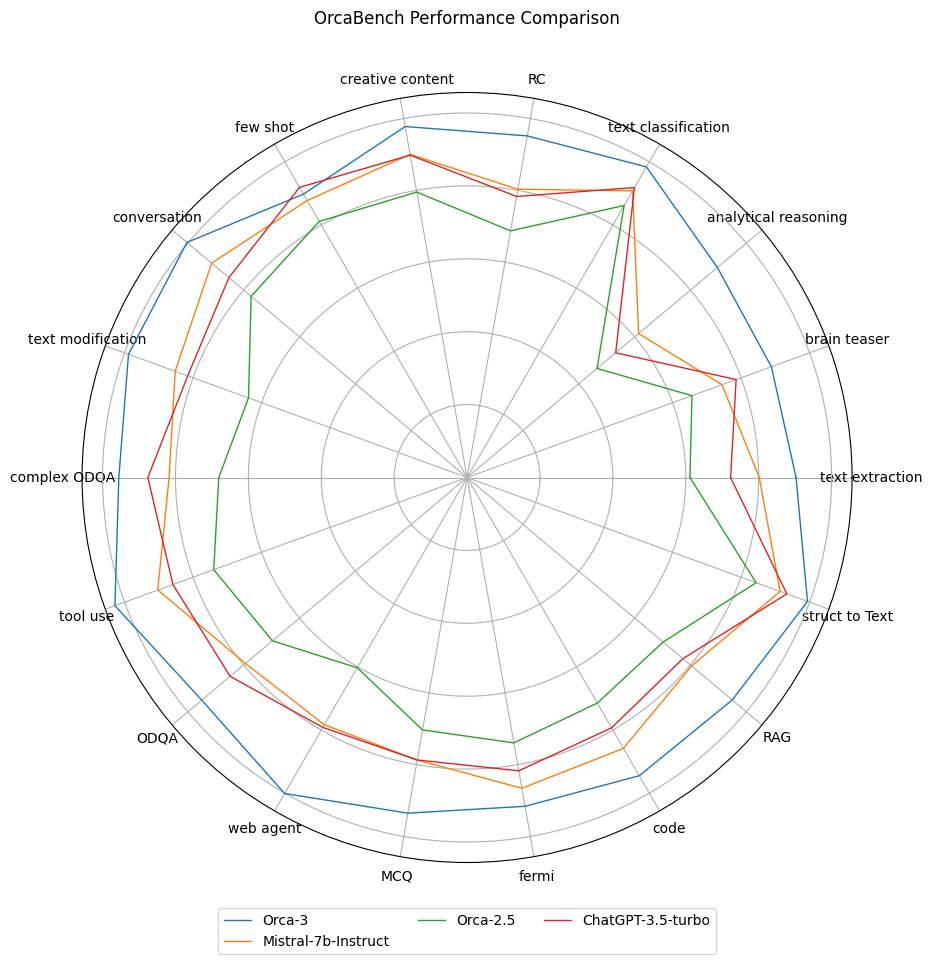}  
    \caption{Performance Comparison between Baselines and Orca3 Checkpoints. Scores are relative to GPT-4 with the outer circle denoting GPT-4's score of 10.}  
    \label{fig:orca-bench-graph}  
\end{figure}  

The Orca-Bench dataset serves as a held-out test set, consisting of 100 samples from each of the 17 skills for which data was curated using AgentInstruct, except for the Open Domain Question Answering (ODQA) category, where we created two test sets. The first subset, referred to as ODQA, consists of 100 questions originated from the initial seed instruction phase. The second subset, termed Complex ODQA, includes more intricate questions developed during the refinement phase.  
   
We evaluated the performance of all baselines using the Orca-Bench dataset. These were scored relative to GPT-4 on a scale ranging from 0 to 10. It is worth noting that some of the entries within Orca-Bench involve multiple exchanges. To illustrate, let a multi-turn interaction in Orca-Bench be denoted by the sequence (system message, $user_1$, assistant, $user_2$, assistant, ...), where each turn is crafted by GPT4 (teacher). For every $user_i$ input, we generate a corresponding student response, which is conditioned on the preceding conversation history as established by the teacher. We then evaluate the student's generated response against the original teacher's response, rating each on a scale from 0 to 10. To calculate a student's overall score, we sum the student's individual scores and divide this total by the sum of the teacher's scores. This ratio is then multiplied by 10 to normalize the student's final score to a 0 to 10 scale.

AgentInstruct's objective is to synthesize a large and diverse corpus of data with varying degrees of difficulty. An efficacious execution  of this strategy should yield a dataset against which baseline models like Orca-2.5, Mistral-Instruct-7b, and ChatGPT score substantially below 10, demonstrating their relative inferiority to GPT-4—a model that is designated as the benchmark with a score of 10. The performance comparison, as depicted in Figure \ref{fig:orca-bench-graph}, illustrates the comparative analysis between baseline models and  Orca-3 . This figure shows the notable enhancement in a broad spectrum of capabilities during post-training, enabled by the AgentInstruct data.  
   
Table \ref{tab:orca-bench-score} encapsulates the average (macro) scores across all assessed dimensions. On average, including orca-3 after each training epoch, the inclusion of AgentInstruct data has led to a performance augmentation of 33.94\% over the Orca 2.5 baseline and an enhancement of 14.92\% over Mistral-Instruct-7B.   
  
\begin{table}[h]  
\centering  
\small  
\begin{tabular}{lc}  
\toprule  
\textbf{Model} & \textbf{Orca-Bench Score} \\  
\midrule  
Orca-2.5 & 7.13 \\  
Mistral-Instruct-7B & 8.31 \\  
ChatGPT & 8.13 \\  
Orca-3 (checkpoint epoch 1) & 9.35 \\  
Orca-3 (checkpoint epoch 2) & 9.49 \\  
Orca-3 & 9.55 \\  
\bottomrule  
\end{tabular}  
\caption{Average Performance of Different Models on Orca-Bench. Scores are computed on a scale of 0 to 10; 10 being the score of GPT4.}  
\label{tab:orca-bench-score}  
\end{table}  
   
\subsection{Benchmark Results}

\begin{table}[h]
\centering
\small
\begin{tabular}{l|lll|lll}
\toprule
\textbf{Model} & \makecell{\textbf{Orca-3}\\\textbf{-7B}} & \makecell{\textbf{Orca-2.5}\\\textbf{-7B}} & \makecell{\textbf{Mistral-}\\\textbf{7B-}\\\textbf{Instruct}}& 
\makecell{\textbf{LLAMA3-}\\\textbf{8B}\\\textbf{instruct}} & 
\makecell{\textbf{GPT-3.5-}\\\textbf{turbo}} & \textbf{GPT-4} \\
\midrule
\textbf{AGIEval} & 56.80 (+40\%) & 42.71 & 40.52 & 47.17 & 50.91 & 61.99 \\
\textbf{MMLU} & 69.95 (+19\%) & 60.34 & 58.61 & 63.44 & 68.26 & 67.07 \\
\textbf{ARC} & 92.47 (+12\%) & 86.39 & 82.72 & 85.74 & 92.0 & 93.35 \\
\textbf{BBH} & 61.83 (+38\%) & 48.63 & 44.71 & 54.97 & 54.17 & 76.06 \\
\textbf{GPQA} & 28.12 (-4\%) & 27.68 & 29.46 & 28.12 & 27.9 & 33.93 \\
\textbf{DROP} & 71.14 (+22\%) & 65.19 & 58.12 & 68.44 & 67.15 & 67.36\\
\textbf{GSM8K} & 83.09 (+54\%) & 74.3 & 54.06 & 77.48 & 78.1$^*$ & 86.88 \\
\midrule
\textbf{FOFO} & 84.01 (+12\%) & 66.19 & 75.3 & 79.35 & 76.92 & 87.45 \\
\textbf{IFEval} & 49.54 (+2\%) & 45.29 & 48.61 & - & 58.6 & 79.3 \\
\textbf{MT-Bench} & 8.20 (+9\%) & 7.15 & 7.53 & 7.99 & 8.01 & 9.04 \\
\textbf{AlpacaEval} & 24.80 (+45\%) & 13.47 & 17.1 & 22.9 & 22.7 & 55 \\
\textbf{InfoBench} & 84.30 (+4\%) & 79.6 & 81 & - & 86.7 & 89.4 \\
\midrule
\multicolumn{7}{l}{\textbf{EQBench}} \\
\textbf{  Metric-v2} & 91.36(+4\%) & 88.03 & 87.75 & 88.67 & 88.95 & 93.32 \\
\textbf{  Metric-v1} & 50.28 (+28\%) & 38.8 & 39.27 & 42.13 & 42.05 & 55.98 \\
\bottomrule
\end{tabular}
\caption{Performance of Orca-3 and other baseline models on all the benchmarks. Note: GPT-3.5-turbo scores for GSM8K are taken from \cite{phi3}. We show in (+x\%) the relative improvement over Mistral-7b-Instruct.}
\label{tab:results_all}
\end{table}

We evaluate Orca-3 against 5 baseline models including Orca-2.5, Mistral-7B-Instruct-v0.3, LLAMA3-8B-Instruct, GPT-3.5-turbo and GPT-4 on the following benchmarks:

\begin{itemize}
    \item \textbf{AGIEval}: AGIEval \cite{zhong2023agieval} is a human-centric benchmark that evaluates a model's abilities in tasks pertinent to human-cognition and problem-solving. It evaluates how well models perform in answering questions from human-centric standardized exams such as SAT, LSAT and math competitions.   
    \item \textbf{MMLU}: Massive Multitask Language Understanding (MMLU) \cite{hendrycks2021measuring} benchmark measures a model's multitask understanding. The benchmark includes approximately 16000 multiple choice questions covering a wide range of 57 academic subjects such as maths, philosphy, medicine, psychology, computer-science, law etc. testing both general and specialized knowledge of the model being tested.
    \item \textbf{ARC}: The AI2 Reasoning Challenge (ARC) \cite{allenai:arc} benchmark, developed by AllenAI, measures the reasoning, commonsense knowledge and deep comprehension abilities of language models. The test set contains 3548 multiple-choice questions that are divided into 2 sets : Easy(2376) and Challenge(1172). 
    \item \textbf{BBH}: Big Bench Hard \cite{suzgun2022challenging} consists of a set of 23 tasks selected from the broader Big-Bench benchmark spanning a wide array of academic subjects requiring complex, multi-step reasoning.
    \item \textbf{GPQA}: Graduate-level Google-Proof Q\&A \cite{rein2023gpqagraduatelevelgoogleproofqa} is a challenging benchmark of 448 high-quality and extremely difficult multiple-choice questions created by domain experts(pursuing PhDs in their domains) in biology, chemistry and physics.
    \item \textbf{DROP}: Discrete Reasoning over Paragraphs \cite{dua-etal-2019-drop} is a Reading Comprehension benchmark requiring the models to resolve references in questions and perform discrete operations over them such as sorting, counting, addition etc.
    \item \textbf{GSM8K}: Grade School Math 8K \cite{cobbe2021training} is a dataset of high quality diverse grade school math word problems. The test split of the dataset consists of 1.32K problems requiring between 2 and 8 steps to solve primarily involving sequence of elementary calculations using basic arithmetic operations.  
    \item \textbf{FoFo}: Format Following \cite{xia2024fofo} is a benchmark that evaluates a model's ability to follow complex, domain-specific formats. The benchmark tests format following on a diverse range of real-world formats and instructions from domains like Healthcare, Finance, Marketing etc. created using AI-Human collaboration.
    \item \textbf{IFEval}: Instruction-Following Evaluation \cite{zhou2023IFEval} is a benchmark measuring a model's ability to follow natural language instructions using a set of 500 prompts covering 25 types of 'verifiable instructions' where each prompt can contain one or more of these instructions.
    \item \textbf{MT-Bench}: MT-Bench \cite{mt-bench2023} benchmark is specifically designed to assess the competence of chat assistants in multi-turn conversations using GPT-4 as the evaluator.
    \item \textbf{AlpacaEval}: AlpacaEval \cite{alpaca_eval} is a benchmark specifically designed for chat-based language models to assess their abilities in the context of instruction-following tasks. It is a single-turn benchmark consisting of 805 instructions representative of user interactions on Alpaca web demo.
    \item \textbf{InFoBench}: The InFoBench \cite{qin2024infobench} benchmark evaluates models instruction following capability using a new metric called Decomposed Requirements Following Ratio(DRFR). DRFR breaks complex instructions down into simpler criteria and facilitates analysis of an LLM's compliance to these decomposed tasks in detail. The benchmark has 500 diverse instructions and 2250 decomposed questions across multiple constraint categories.
    \item \textbf{EQBench}: This Emotional Intelligence benchmark \cite{paech2024eqbench} evaluates aspects of emotional intelligence in language models. It tests models capabilities to comprehend intricate emotions and social interactions by providing a conversation between characters and then asking the model to predict intensity of emotional states of those characters. The authors discovered a strong correlation (r=0.97) between EQ-Bench and comprehensive multi-domain benchmarks like MMLU. 
\end{itemize}
The results for all the baselines on each benchmark are given in table \ref{tab:results_all}. All of the evaluations for Orca-3 and other baselines was done in a zero-shot setting unless mentioned otherwise in the text.

 The types of tasks/benchmarks and the corresponding method used to extract answer and generate metrics is specified in Appendix~\ref{sec:eval_details}.

\subsection{Evaluation: Reading Comprehension}

\begin{table}[!htb]
\centering
\small
\begin{tabular}{m{3cm}|m{2.5cm}m{1.5cm}m{1.8cm}|m{1.4cm}m{1.2cm}}
\toprule
\textbf{Model} & \makecell{\textbf{Orca-3}\\\textbf{-7B}} & \makecell{\textbf{Orca-2.5}\\\textbf{-7B}} &
\makecell{\textbf{Mistral-}\\\textbf{7B-Instruct}} & \makecell{\textbf{GPT-3.5-}\\\textbf{turbo}} & \textbf{GPT-4} \\
\midrule
\textbf{AGIEval lsat-rc} & 75.84 (+21\%,+20\%) & 62.45 & 63.2 & 63.57 & 72.86 \\
\textbf{AGIEval sat-en} & 87.38 (+13\%,+15\%) & 77.18 & 75.73 & 82.04 & 82.52 \\
\makecell[l]{\textbf{AGIEval}\\\textbf{gaokao-english}} & 87.25 (+13\%,+17\%) & 77.45 & 74.84 & 83.01 & 87.25 \\
\textbf{AGIEval lsat-lr} & 63.14 (+45\%,+36\%) & 43.53 & 46.27 & 54.9 & 68.82 \\
\midrule
\textbf{DROP} & 71.14 (+9\%,+22\%) & 65.19 & 58.12 & 67.15 & 67.36 \\
\midrule
\textbf{Average} & 76.95 (+18\%,+21\%) & 65.16 & 63.63 & 70.13 & 75.76 \\
\bottomrule
\end{tabular}
\caption{Performance of models on reading comprehension based sub-tasks and benchmarks. The figures (x\%, y\%) adjacent to the Orca-3 results signify the percentage of improvement compared to Orca 2.5 and Mistral-7B-Instruct, respectively.}
\label{tab:results_reading_comprehension}
\end{table}

Reading comprehension is a crucial capability for LLMs. It is arguably even more important for Small Language Models (SLMs), as they are better suited as reasoning engines than mere retrieval systems. Through targeted training with AgentInstruct, we observe substantial improvement in Mistral's reading comprehension capabilities (Table \ref{results_reading_comprehension})—showcasing an  18\% improvement over Orca 2.5 and a 21\% gain relative to Mistral-Instruct-7b. Furthermore, by leveraging this data-driven approach, we have elevated the performance of a 7B model to match that of GPT-4 on the reading comprehension sections of the Law School Admission Tests (LSATs), which are considered difficult for human test-takers.

\subsection{Evaluation: Math}

\begin{table}[!htb]
\centering
\small
\begin{tabular}{m{3.1cm}|m{2.5cm}m{1.5cm}m{1.8cm}|m{1.4cm}m{1.2cm}}
\toprule
\textbf{Model} & \makecell{\textbf{Orca-3}\\\textbf{-7B}} & \makecell{\textbf{Orca-2.5}\\\textbf{-7B}} & \makecell{\textbf{Mistral-}\\\textbf{7B-Instruct}} & \makecell{\textbf{GPT-3.5-}\\\textbf{turbo}} & \textbf{GPT-4} \\
\midrule
\makecell[l]{\textbf{AGIEval math}} & 42.90 (+73\%,+168\%) & 24.8 & 16.0 & 38.0 & 57.9 \\
\makecell[l]{\textbf{AGIEval sat-math}} & 80.91 (+34\%,+50\%) & 60.45 & 54.09 & 67.73 & 90.0 \\
\midrule
\makecell[l]{\textbf{BBH multistep}\\\textbf{-arithmetic-two}} & 66.80 (+1418\%,+882\%) & 4.4 & 6.8 & 46.4 & 77.2 \\
\midrule
\makecell[l]{\textbf{MMLU abstract}\\\textbf{algebra}} & 55.00 (+129\%,+104\%) & 24.0 & 27.0 & 47.0 & 70.0 \\
\makecell[l]{\textbf{MMLU college}\\\textbf{mathematics}} & 44.00 (+63\%,+44\%) & 30.0 & 34.0 & 39.0 & 62.0 \\
\makecell[l]{\textbf{MMLU high-school}\\\textbf{mathematics}} & 66.67 (+41\%,+94\%) & 47.41 & 34.44 & 57.04 & 66.67 \\
\midrule
\makecell[l]{\textbf{GSM8K}} & 83.09 (+12\%,+54\%) & 74.3 & 54.06 & 78.1$^*$ & 86.88 \\
\bottomrule
\end{tabular}
\caption{Performance scores of models on Math benchmarks. Note: GPT-3.5-turbo accuracy scores reported for GSM8K are taken from Phi3 paper\cite{phi3}. The figures (x\%, y\%) adjacent to the Orca-3 results signify the percentage of improvement compared to Orca 2.5 and Mistral-7B-Instruct, respectively.}
\label{tab:results_maths}
\end{table}

Assessing the reasoning capabilities of AI models can be effectively accomplished through math problem solving. While SLMs have shown considerable improvement in elementary math, their performance typically falters with more complex high school and college-level mathematics. Math problems are generated by the Open Domain Question Answering and Multiple-Choice Questions Flows. With AgentInstruct, we have managed to enhance Mistral's proficiency across a spectrum of difficulties, ranging from elementary to college-level math (Table \ref{tab:results_maths}. This has led to a signficant  performance boost, with improvements ranging from 44\% to 168\% on various popular mathematical benchmarks. It should be emphasized that the objective of Generative Teaching is to teach a skill than generating data to meet a specific benchmark. The effectiveness of AgentInstruct for Generative Teaching is evidenced by marked enhancements across a variety of mathematical datasets.

\subsection{Evaluation: Format Following}

\begin{table}[h]
\centering
\small
\begin{tabular}{lll}
\toprule
& \textbf{Model} & \textbf{FoFo} \\
\midrule
\midrule
\multirow{5}{*}{Open-source} \\
& \textbf{Orca-3-7B} & 84.01 (+26.92\%,+11.5\%) \\
& \textbf{Orca-2.5-7B} & 66.19 \\
& \textbf{Mistral-7B-Instruct} & 75.3\\
\midrule
\multirow{5}{*}{Closed-source} \\
& \textbf{GPT-3.5-turbo} & 76.92\\
& \textbf{Gemini Pro} & 80.25$^*$ \\
& \textbf{GPT-4} & 87.45\\
\bottomrule
\end{tabular}
\caption{Performance of Orca-3-7B model and other open and closed-source baselines on FoFo benchmark. The figures (x\%, y\%) adjacent to the Orca-3 results signify the percentage of improvement compared to Orca 2.5 and Mistral-7B-Instruct, respectively. Note: The scores for Gemini Pro are taken from the original paper \cite{xia2024fofo}} 
\label{tab:results_FoFo}
\end{table}

Following formatting guidelines is essential for language models to be applicable in real-world situations. In all AgentInstruct flows, we ensure that format-following is taught for each particular scenario, by synthesizing, through agents, several formatting guidelines. By doing so, we are able to significantly improve (11.5\%) Mistral's ability to follow formats, surpassing even the capabilities of Gemini Pro.

\subsection{Evaluation: Abstractive Summarization}

Summarization is an important capability for Language Models, with many models achieving high quality summarization performance, yet struggling with hallucination. We assessed  summarization ability using two key metrics: hallucinations and quality. For this purpose, we utilized GPT4 as our evaluator. The prompts utilized in these evaluations can be found in Appendix~\ref{sec:eval_details}.  We used the following benchmarks for evaluating summarization abilities: 

\begin{itemize}
        \item \textbf{ACI-Bench:} The Ambient Clinical Intelligence Benchmark (ACI-Bench)~\cite{yim2023acibench} is a dataset designed for benchmarking automatic report generation from doctor-patient conversations. The test set comprises 120 data points.
        
        \item \textbf{InstruSum:} A dataset~\cite{liu2023benchmarkinggenerationevaluationcapabilities} for evaluating the generation capabilities LLMs for instruction-controllable summarization. It consists of 100 datapoints.
    
    \item \textbf{Orca-Sum:} A newly created benchmark to evaluate LLMs' ability to follow summarization and grounded data transformation instructions. To construct this test set, we sampled data from 45 summarization datasets collected from Hugging Face across multiple domains such as news, conversations, science, health, social, e-mails, code, etc. for a total of 458 datapoints. We randomly collected, up to 1000 datapoints which then we carefully deduplicated to avoid overlapping with the training set. We then used GPT-4 to generate a set of 40 prompts for each dataset out of each we randomly sampled one for each selected datapoint. The prompts are dataset-specific and focus on summarization, grounding, and data transformation. For instance, a prompt may ask the model to generate a TikTok video out of a scientific paper or a legal contract from a Wikipedia page. This allows us to measure not only the quality of the response but also hallucination in a challenging scenario, as the model is forced to move between formats and domains.
\end{itemize}

\begin{table}[h]
\centering
\small
\begin{tabular}{lp{42px}ccccc}
\toprule
\textbf{Model} & \makecell{\textbf{Orca-3}\\\textbf{-7B}} & \makecell{\textbf{Orca-2.5}\\\textbf{-7B}} & \makecell{\textbf{Mistral-}\\\textbf{7B-Instruct}} & \makecell{\textbf{LLAMA3-}\\\textbf{8B-instruct}}  & \makecell{\textbf{GPT-3.5-}\\\textbf{turbo}} & \textbf{GPT-4} \\
\midrule
\multicolumn{7}{c}{\textbf{Hallucination Rate (\%) - Smaller is better}} \\
\midrule
\textbf{All (micro)} & 21.09 (-26.12\%, -31.34\%) & 28.55 & 30.72  & 34.22 & 21.13 & 15.07 \\
\textbf{Orca-Sum} & 28.17 & 36.84 & 39.61  & 38.43 & 28.60 & 21.66 \\
\textbf{InstruSum} & 9.00 & 12.00 & 17.00  & 25.00 & 12.00 & 1.00 \\
\textbf{ACI-Bench} & 4.20 & 10.83 & 8.30  & 25.83 & 1.70 & 1.70 \\
\midrule
\multicolumn{7}{c}{\textbf{Quality Score (1-10) - Higher is better}} \\
\midrule
\textbf{All (micro)} & 9.14 ~~(+7.91\%, +3.28\%) & 8.47 & 8.85 & 9.00 & 8.69 & 9.08 \\
\textbf{Orca-Sum} & 8.95 & 8.27 & 8.61  & 8.90 & 8.32 & 8.61 \\
\textbf{InstruSum} & 9.17 & 8.30 & 9.16  & 9.21 & 9.27 & 9.31 \\
\textbf{ACI-Bench} & 9.72 & 9.39 & 9.48  & 9.23 & 9.60 & 9.70 \\
\bottomrule
\end{tabular}
\caption{Hallucination rates and quality scores evaluated by GPT4. The figures (x\%, y\%) adjacent to the Orca-3 results signify the percentage of improvement compared to Orca v2.5 and Mistral-7B-Instruct, respectively.}
\label{tab:results_summarization}
\end{table}
The results are presented in Table \ref{tab:results_summarization}. With the AgentInstruct approach, we successfully achieved a reduction in hallucinations by 31.34\%, while attaining a quality level comparable to GPT4 (Teacher).

\subsection{Evaluation: RAG}
\begin{table}[!htbp]
\small
\begin{tabular}{m{2.2cm}m{0.7cm}m{1.3cm}m{1.3cm}m{1.4cm}m{1.4cm}m{1.3cm}m{1.3cm}}
\toprule
                            &     & \multicolumn{6}{c}{MIRAGE Datasets}                     \\      \cmidrule{3-7}
                            &     & MMLU-Med & MedQA-US & MedMCQA & PubMedQA  & BioASQ     & Avg.  \\ \midrule
\multirow{2}{*}{\makecell[l]{\textbf{GPT-4 }\\\textbf{(0613)}}} & CoT & 89.44    & 83.97    & 69.88   & 39.6      & 84.3       & 73.44 \\
 & RAG & 87.24    & 82.8     & 66.65   & 70.6      & 92.56      & 79.97 \\ \midrule
\multirow{2}{*}{\makecell[l]{\textbf{GPT-3.5-turbo }\\\textbf{(0613)}}}  & CoT & 72.91    & 65.04    & 55.25   & 36        & 74.27      & 60.69 \\
  & RAG & 75.48    & 66.61    & 58.04   & 67.4      & 90.29      & 71.57 \\ \midrule
\multirow{2}{*}{\makecell[l]{\textbf{Orca-2.5-7B}}} & CoT & 63.91    & 51.37    & 43.65   & 29.6      & 71.04      & 51.92 \\
                            & RAG & 53.72    & 37.08    & 39.23   & 19        & 69.09      & 43.62 \\ \midrule
\multirow{2}{*}{\makecell[l]{\textbf{Mistral-7B-}\\\textbf{Instruct-v0.1}}} & CoT & 50.96    & 42.73    & 34.9    & 27.6      & 47.57      & 40.75 \\
 & RAG & 54.64    & 35.35    & 43.41   & 30.2      & 68.77      & 46.47 \\ \midrule\\
 \multirow{2}{*}{\makecell[l]{\textbf{Orca-3-7B}}} & CoT & 71.35    & 55.38    & 51.33   & 27.8      & 75.24      & 56.22 \\
 & RAG & \makecell[l]{71.17 \\(+30.25\%)}    & \makecell[l]{51.85\\(+46.68\%)}    & \makecell[l]{57.95\\(+33.49\%)}  & \makecell[l]{58.2\\ (+92.71\%)}      & \makecell[l]{82.2\\(+19.52\%)}       & \makecell[l]{64.27\\(+38.30\%)} \\ 
\bottomrule
\end{tabular}
\caption{Evaluation results of RAG skill on MIRAGE. The figures (x\%) adjacent to the Orca-3 results signify the percentage of improvement compared to Mistral-7B-Instruct respectively. CoT shows the performance of the same models when answering directly without using RAG}
\label{tab:mirage}
\end{table}
The RAG (Retrieval Augmented Generation) skill significantly boosts the capacity of Language Models to generate informed, contextually precise responses, hence upgrading their overall performance and usefulness. It is arguably more effective to test the RAG proficiency of language models in areas where the models have limited knowledge. For this study, we selected MIRAGE\cite{xiong2024benchmarking}, a benchmark that focuses on answering medical questions by referring to information retrieved from a medical corpus. Since the medical domain is not typically a primary focus of the models evaluated in this study, MIRAGE provides an effective platform for assessing their RAG capabilities. Additionally, AgentInstruct RAG data used generic, non medical data seed, enabling us to test how well can the skill (RAG) be applied to new domains. 

We use the same retrieval mechanism across all models on MIRAGE~\cite{xiong2024benchmarking}, using MedRAG, the corpus accompanying the benchmark. This involves using the same retrieval function and the same number of retrieved documents for all models. As all models are presented with the same set of retrieved documents, the comparison accurately reflects the ability of different models to incorporate retrieved results into their responses.

Table \ref{tab:mirage} shows the results of all models on MIRAGE with and without leveraging RAG\footnote{Results of GPT-4 and GPT-3.5-Turbo are from~\cite{xiong2024benchmarking}.}. Overall, we observe that
\begin{itemize}
    \item Models with a deeper understanding of the task (CoT scores) tend to have higher RAG scores. If we restrict our focus to only RAG performance, applying post-training, we've managed to enhance Mistral's performance by an average of 38.30\%.
    \item Of the five datasets in MIRAGE, PubMedQA arguably offers the most effective testbed for assessing models ability to do RAG. In PubMedQA, all models have limited prior knowledge, and the retrieved context provides essential information, as demonstrated by GPT4's performance leap. All Mistral fine-tunes exhibit similar performance, but only Orca-3 (Mistral trained with AgentInstruct RAG flow data) shows a substantial improvement, resulting in a relative improvement of 92.71\% over Mistral-Instruct.
\end{itemize}

\label{subsubsec:safety_datasets}
\section{Limitations}
\label{sec:limitations}
AgentInstruct reduces human expertise required for data generation significantly and enables creating of high-quality synthetic data at scale. However, this is till an early step in this direction and could suffer from many limitations associated with synthetic data generation, including but not limited to:

\noindent {\bf Extensibility:} Creating the agentic flows for different skills  depends on human effort for the construction of the flows. Future work should consider how to  automate the construction of the agentic flow from the user specification.

\noindent {\bf Accuracy:} Synthetic data may not perfectly replicate the complexity and nuances of real-world data, leading to potential inaccuracies. Additional work is needed to better assess the quality of the data.

\noindent {\bf Cost}: Generating synthetic data with multiple agents using LLMs and tools can be resource-intensive.

\noindent {\bf Bias:} If the original seed data used to generate synthetic data contains biases, these biases can be reflected and even amplified in the synthetic data.

\noindent {\bf Validation}: It can be difficult to validate synthetic data to ensure it accurately represents the desired scenarios.

\noindent {\bf Dependency on Seed Data}: The quality of synthetic data is dependent on the quality of the real data used as seeds. Poor quality input data could  result in poor quality synthetic data.

\sysname{} is fine-tuned with the AgentInstruct data based on the Mistral  model family, and retains many of its limitations, as well as the common limitations of other large language models and limitations originating from its training process, including:

\noindent {\bf Data Biases:} Large language models, trained on extensive data, can inadvertently carry biases present in the source data. Consequently, the models may generate outputs that could be potentially biased or unfair.

\noindent {\bf Lack of Transparency:} Due to the complexity and size, large language models can act as ``black boxes'', making it difficult to comprehend the rationale behind specific outputs or decisions. We recommend reviewing transparency notes from Azure for more information\footnote{\url{https://learn.microsoft.com/en-us/legal/cognitive-services/openai/transparency-note}}.

\noindent {\bf Content Harms:} There are various types of content harms that large language models can cause. It is important to be aware of them when using these models, and to take actions to prevent them. It is recommended to leverage various content moderation services provided by different companies and institutions. On an important note, we hope for better regulations and standards from government and technology leaders around content harms for AI technologies in future. We value and acknowledge the important role that research and open source community can play in this direction.

\noindent {\bf Hallucination:} It is important to be aware and cautious not to entirely rely on a given language model for critical decisions or information that might have deep impact as it is not obvious how to prevent these models from fabricating content. Moreover, it is not clear whether small models may be more susceptible to hallucination in ungrounded generation use cases due to their smaller sizes and hence reduced memorization capacities. This is an active research topic and we hope there will be more rigorous measurement, understanding and mitigations around this topic. 

\noindent {\bf Potential for Misuse:} Without suitable safeguards, there is a risk that these models could be maliciously used for generating disinformation or harmful content.

\noindent {\bf Data Distribution:} \sysname{}'s performance is likely to correlate strongly with the distribution of the tuning data. This correlation might limit its accuracy in areas underrepresented in the training dataset.
\section{Conclusions}
The AgentInstruct approach to Generative Teaching offers a promising solution to the challenge of generating large amount of diverse and high-quality data for model post-training. This method stands out by using agentic flows for synthetic data generation, thus addressing key concerns associated with the use of synthetic data in model training, such as the lack of diversity and the need for intensive human curation and intervention during the data creation process. By leveraging an agentic framework, AgentInstruct can generate tailored datasets comprising both prompts and responses from unstructured data sources, facilitating the post-training of models and teaching them variety of skills. The efficacy of this approach is exemplified by the substantial improvement observed in the \sysname{} model, which, post-trained with a 25M pair dataset generated by AgentInstruct, showcased a notable performance gain across multiple benchmarks. We believe using agentic flows for creating synthetic data can show significant value for all stages of model training, including pre-training, post-training and domain/task specialization. The ability to use unstructured content to generate diverse and high-quality instruction data given any specifications could pave the way for creating (semi) automated pipelines using synthetic data for model customization (using domain specific content as seeds) and continual improvement (generating higher quality data than the base model with agentic flows).

\bibliographystyle{plainnat} 
{
\small
\bibliography{anthology,custom}
}
\clearpage
\appendix

\section{Agentic Flows Details}
\label{sec:appendix-additional-details}

\subsection{Reading Comprehension Flow}
\textbf{Reading Comprehension transformation agents}:  
  
\begin{enumerate}  
  \item \textbf{Argument Passage Generator:} This agent is adept at creating passages that articulate arguments, which may occasionally contain logical inconsistencies.  
  \item \textbf{Debate Passage Generator:} It specializes in crafting passages that mimic the structure and content of debate transcripts.  
  \item \textbf{Conversation Passage Generator:} This agent generates passages that depict dialogues.  
  \item \textbf{Meeting Transcript Generator:} It is designed to produce meeting transcripts.  
  \item \textbf{Poem Generator:} This agent generates poems.  
  \item \textbf{Satirical Passage Generator:} It creates texts infused with satirical wit.  
  \item \textbf{Instructional Passage Generator:} This agent generates passages resembling instructional manuals.  
  \item \textbf{Long Text Generator:} It extends the original text by incorporating additional information, thereby increasing its length.  
  \item \textbf{Identity Agent:} A straightforward agent that replicates the input text verbatim.  
\end{enumerate}  

\textbf{Instruction Taxonomy for Seed Instruction Generation Flow}

\begin{enumerate}  
    \item Literal Comprehension Question (Short Answer(or list)): a question that asks for a specific detail(s) or fact(s) clearly stated in the text.    
    \item Numerical Discrete Reasoning (Reasoning):  questions that require the reader to use numerical reasoning over many facts from the text.  
    \item Critical Comprehension Question (True/False): construct two statements about the purpose or point of view that the reader must assess as true or false, with one being true and the other false.  
    \item Evaluative Comprehension Question (Essay): an open-ended question that prompts an in-depth analysis of the text's theme or the effectiveness of an argument.  
    \item Vocabulary and Language Use (Fill-in-the-Blank): a fill-in-the-blank question that tests understanding of a particular word or phrase used in the text.  
    \item Relationship Comprehension Question (Matching): a matching question where respondents pair items  based on a specific criterion.  
    \item Sequencing Events (Ordering):  a series of events from the text arranged in the correct chronological order.  
    \item Strengthen: identify information that would make the argument's conclusion more likely to be true.  
    \item Weaken: find evidence or an argument that would make the conclusion less likely to be true.  
    \item Assumption (Necessary Assumption): determine what must be true for the argument to hold.  
    \item Flaw: point out a mistake in the argument's reasoning.  
    \item Inference (Must Be True): Choose an option that logically follows from the information provided.  
    \item Principle (Identify the Principle): Recognize the general rule or principle that underlies the argument.  
    \item Method of Reasoning (Describe the Argument): Describe how the argument is constructed logically.  
    \item Resolve the Paradox: Offer an explanation that reconciles seemingly contradictory information.  
\end{enumerate}

\subsection{Text Modification Flow}

\textbf{Instruction Taxonomy for Seed Instruction Generation Flow}  
  
\begin{enumerate}  
    \item Paraphrasing: Rewriting text using different words and sentence structures while maintaining the original meaning.  
    \item Text Simplification: Making text easier to read and understand by using simpler words and sentence structures, often for children or language learners.  
    \item Text Expansion: Adding more information or detail to make text more comprehensive or to meet a certain word count.  
    \item Text Translation: Converting text from one language to another while attempting to preserve the original meaning as closely as possible.  
    \item Text Formatting: Altering the appearance of text to improve readability or for stylistic purposes.  
    \item Sentiment Modification: Changing the tone of the text to alter its emotional impact, such as making a sentence sound more positive or negative.  
    \item Text Annotation: Adding notes, comments, or explanations to a text, often for the purpose of analysis or to provide additional context.  
    \item Keyword Replacement: Substituting specific words or phrases with synonyms or related terms.  
    \item Text Removing: Redacting or removing content from text.  
    \item Text Capitalization: Adjusting the case of letters in text, such as converting to uppercase, lowercase, title case, or sentence case, starting every sentence with a particular letter, word.  
    \item Text Styling: Applying styles like bold, italics, underline, etc., to emphasize certain parts of the text or for aesthetic purposes.  
    \item Content Rewriting: Extensively modifying a text to produce a new version, which could involve changing the perspective, style, or target audience.
    \item Data Normalization: Standardizing text to ensure consistency, such as converting dates and times to a standard format or unifying the spelling of words.  
    \item Plagiarism Rewording: Altering text to avoid plagiarism, ensuring that the content is original.  
    \item Code Switching: Alternating between languages or dialects within a text, often to reflect bilingual speakers' patterns or for creative writing.  
    \item Text Obfuscation: Intentionally making text vague or harder to understand, sometimes for security purposes (like masking personal data).  
    \item Textual Entailment: Modifying a sentence or phrase to either entail or contradict another sentence, often used in natural language processing tasks.  
    \item Rewriting with vocabulary limitations: Rewriting the entire text or a piece of it while using a limited vocabulary. For example, all words should start with letter 'a', all n-th word should start with letter 'b', each sentence should start with a 'vowel', etc.  
\end{enumerate}  
\section{Evaluation Details}
\label{sec:eval_details}
 
 The types of tasks/benchmarks and the corresponding method used to extract answer and generate metrics is specified below:

\begin{itemize}
    \item \textbf{Multiple Choice Questions}: All the models are evaluated in an open-ended generation setting with an empty system message We then use GPT-4 for extraction of the option selected by the model from model's response instead of regex based extraction done in \cite{mitra2023orca2teachingsmall}. The extracted prediction is matched with the ground truth to generate accuracy scores.

    The system message used for the GPT-4 extractions is as follows:
    \begin{mdframed}
        \fbox{MCQ GPT-4 Extraction System Message}
        
        You are an Evaluator Assistant. 
        You support the exam evaluator by parsing student responses.
        You are an unbiased Evaluator and do not rely on your knowledge 
        but stick to the user provided context.
        You are provided with the question, 
        answer options and a student's response.
        Your task is to parse the option student selected in their response as their final answer 
        and return the alphabet ID of that answer in the provided options.
        If the student gave multiple answers return them as a list. 
        
        Use the following format:
        
        Parsed Student Answer: Final answer extracted from Student's response. This should only be the alphabets representing the option the student chose.  
        
        Example 1:
        
        Input :
        
        Question:
        
        Find all c in $Z_3$ such that $Z_3[x]/(x^2 + c)$ is a field.
        
        Student Response : 
        
        I think 0 is incorrect, so is 2. 
        3 seems incorrect as well. 
        I think 1 is the correct final answer.
        
        Options :
        
        [(A) 0, (B) 1, (C) 2, (D) 3 ]
        
        Output:
        
        Parsed Student Answer: B

        Example 2:
        
        Input :
        
        Question: 
        
     Find all c in $Z_3$ such that $Z_3[x]/(x^2 + c)$ is a field.
        
        Student Response : 
        
        I think 0 is incorrect. 
        3 seems incorrect as well. 
        I think 1 and 2 could be the correct final answers.
        
        Options :
        
        [(A) 0, (B) 1, (C) 2, (D) 3 ]
        
        Output:
        
        Parsed Student Answer: [B,C]
 
    \end{mdframed}

    \item \textbf{Exact Match/Span Extraction Problems}: For tasks with math based questions like GSM8K and problems where a ground-truth answer value is given (like DROP), we prompt the models being evaluated to generate the answer and use GPT-4 to extract the exact answer and also match it with the ground-truth provided to produce a final verdict of whether the model's answer was 'Correct' or 'Incorrect'. We use a specific system message for maths based questions, and another for all the other exact match/span extraction problems, both of which are provided below.

    \begin{mdframed}
        \fbox{Maths GPT-4 Extraction System Message}

        As an expert Math teacher, your role is to evaluate a student's answer to a word problem. 
        The problem is accompanied by a correct solution provided by the problem setter. 
        It is important to remember that there may be various methods to solve a word problem, 
        so the student's steps might not always align with those in the problem setter's solution. 
        However, the final answer, typically a number, 
        should be unique and match the problem setter's answer. 
 
        Use the following format:
 
        Error Analysis:
        
        In one sentence, extract the final answer from the problem setter's solution and compare it with the student's answer. Do they match?
        
        Final Verdict:
        
        Correct/Incorrect

        \fbox{General Extraction System Message}
        
        You are an Evaluator Assistant. 
        You support the exam evaluator by parsing student responses.
        You are an unbiased Evaluator and 
        do not rely on your knowledge but stick to the user provided context.
        You are provided with the correct answer and a student's response.
        Your task is to parse the answer from student's response 
        and then match it with the correct answer.
        If the student's final answer matches the correct answer provided, 
        output a 'Correct', else an 'Incorrect'.
        
        Please rely strictly on the correct answer given in the context only.
        
        Use the following format:
 
        Error Analysis:
        
        In one sentence, extract the final answer from the student's solution and compare it with the correct answer. Do they match?
 
        Final Verdict:
        
        Correct/Incorrect
    \end{mdframed}

    \item \textbf{EQBench}: For EQBench, we prompt the models to generate the emotion scores given the conversation in the prompt and then use GPT-4 to extract the scores generated by the model for each emotion in the prompt. The metric scores are generated using both the version 1 and 2 implementations described in the EQBench paper and the creators' github repository. The scoring calculation is calibrated such that a score of 0 corresponds to answering randomly, and a 100 would denote perfect alignment with the reference answer. The system message used for extraction of emotion scores from evaluated model's response using GPT-4 is given below:
    \begin{mdframed}
        \fbox{EQBench GPT-4 Extraction System Message}
        
        You are a helpful assistant. 
        You will be given a student agent response which will consist of possible emotions and a score from 0-10 for each of those emotions, 
        followed by a step by step critique and then revised scores in the following format, 
        First pass scores:
        
        Emotion1: <score>
        
        Emotion2: <score>
        
        Emotion3: <score>
        
        Emotion4: <score>
        
        Critique: <your critique here>
        
        Revised scores:
        
        Emotion1: <revised score>
        
        Emotion2: <revised score>
        
        Emotion3: <revised score>
        
        Emotion4: <revised score>
        
        [End of answer]
        
        Remember: zero is a valid score as well.
        
        You will also be provided with the Emotions. 
        Your task is to parse the Revised scores for each of the emotions from the student agent response.
        Return the revised scores in the student agent response for the emotions in the following format:
        
        {
        
        "Emotion1" : "Score",
        
        "Emotion2" : "Score",
        
        "Emotion3" : "Score",
        
        "Emotion4" : "Score"
        
        }
        
        For example:
        
        Input
        
        Student Agent Response:
        
        First pass scores:
        
        Resigned: 8
        
        Angry: 2
        
        Hopeful: 4
        
        Embarrassed: 9
        
        Critique: 
        
        Elliot is likely to feel resigned because he has just confessed his feelings to Alex, knowing that Alex is already in a relationship. He might feel a bit angry at himself for putting himself in this situation. There is a slight sense of hopefulness in his confession, hoping that Alex might reciprocate his feelings. He is also likely to feel embarrassed for putting Alex in an awkward position.
        
        Revised scores:
        
        Resigned: 7
        
        Angry: 3
        
        Hopeful: 5
        
        Embarrassed: 8
        
        Emotions: 
        
        1. Resigned, 2. Angry, 3. Hopeful, 4. Embarrassed
        
        Output
        
        {
        
        "Resigned" : 7,
        
        "Angry" : 3,
        
        "Hopeful" : 5, 
        
        "Embarrassed" : 8 
        
        }
    \end{mdframed}

    \item \textbf{Open-Ended Generation}: These are the tasks where model is prompted to generate an answer to an open-ended question, but a ground-truth to match the answer is not available. The metric calculation method for the benchmarks in this category are provided below:  
    \begin{itemize}
        \item \textbf{FOFO}: For this benchmark the evaluation is done using a judge, GPT-4(version 0613). We use the judge system message provided in the original paper of the benchmark \cite{xia2024fofo}. GPT-4 is used to give a format correctness score between 0 and 1, 1 meaning the model's response strictly follows the format specified in the prompt and 0 otherwise. The final score is measured as the percentage of times the model being evaluated followed the format specified in the prompt strictly.

        \item \textbf{IFEval}: IFEval benchmark requires checking if the model response follows the verifiable instructions given in the prompt. For this we use the code provided by the authors \cite{zhou2023IFEval}.

        \item \textbf{MT-Bench}: MT-Bench benchmark consists of a first-turn query and a second-turn query independent of the evaluated model's response. The benchmark employs GPT-4 to judge each turn's response and provide a score from 1 to 10. The average score over all interactions is reported. System message and prompt template used is the one provided by the creators \cite{mt-bench2023}.

        \item \textbf{AlpacaEval}: In this benchmark we measure win-rates, i.e. the number of times a powerful LLM (GPT-4-turbo version 0613 in our case) prefers the outputs of the evaluated model over a reference answer \cite{alpaca_eval}.

        \item \textbf{InfoBench}: InfoBench is also evaluated using GPT-4 (version 1106-preview) as the judge determining if the model response follows the decomposed instruction and we use the implementation provided by the creators of the benchmark \cite{qin2024infobench}. 
    \end{itemize}
\end{itemize}

\subsection{Summarization Quality and Hallucination Evaluation}

We use GPT-4 with the following prompts for evaluating quality and hallucination in summarization:

\begin{figure}
\begin{mdframed}
    \fbox{\textbf{Hallucination Judge Example}}
    \begin{verbatim}
    You will be given a summary instruction and a generated summary.
    Your task to decide if there is any hallucination in the generated 
    summary.
    
    User Message:
    {{place summary task here}}
    
    Generated Summary:
    {{place response here}}
    
    =========================
    
    Go through each section in the generated summary, do the following:
    
    - Extract relevant facts from the article that can be used to verify 
    the correctness of the summary
    - Decide if any section contains hallucination or not.
    
    At the end output a JSON with the format:
    
    {"hallucination_detected": "yes/no", "hallucinated_span": "If yes, 
    the exact span of every hallucinated text part from the summary in 
    list format; if no, leave this empty."}
    
    Use the format:
    Analysis:
    section 1:
    write the part of the summary
    relevant segments:
    extract relevant segments from the article
    judgement:
    decide if the section of the summary is supported by the article
    repeat this for all sections
    
    ....
    
    Final verdict:
    {"hallucination_detected": "yes/no", "hallucinated_span": "If yes, 
    the exact span of every hallucinated text part in list format; if no, 
    leave this empty."}
    \end{verbatim}
    
\end{mdframed}
\caption{Prompt template used for hallucination detection in Text Summarization.}
\label{example:hallucination_judge}
\end{figure}

\begin{figure}
\begin{mdframed}
    \fbox{Quality Judge Example}
    \begin{verbatim}
    Please act as an impartial judge and evaluate the quality of the 
    response provided by an AI assistant to the user instruction 
    displayed below.

    Your evaluation should assess the following criteria:

    - Instruction Adherence: Does the response correctly follow the user 
    instruction?
    - Content Grounding: Is the answer grounded in the instruction 
    without introducing new content beyond what is already present? 
    Penalize hallucinations.
    - Overall Quality: Assess the clarity, coherence, and completeness 
    of the response.

    Begin your evaluation with a short explanation highlighting the pros 
    and cons of the answer. Be as objective as possible. After providing 
    your explanation, rate the overall quality of the response on a scale 
    of 1 to 10 using this format: 
    "Rating: [[rating]]" (e.g., "Rating: [[5]]").

    User Instruction:
    {{place instruction here}}
    
    Assistant's Response:
    [The Start of Assistant's Answer]
    
    {{place response here}}
    
    [The End of Assistant's Answer]
    \end{verbatim}
\end{mdframed}
\caption{Prompt template for evaluation of summary quality.}
\label{example:quality_judge}
\end{figure}
\end{document}